\documentclass[10pt,twocolumn,letterpaper]{article}
\pdfoutput=1 

\usepackage{3dv}
\usepackage{times}
\usepackage{epsfig}
\usepackage{graphicx}
\usepackage{amsmath}
\usepackage{amssymb}
\usepackage{amsmath,amssymb} 
\usepackage{mathptmx}
\usepackage{graphicx}
\usepackage{multirow}
\usepackage{times}
\usepackage{color}
\newcommand{\todo}[1]{}
\renewcommand{\todo}[1]{{\color{red} TODO: {#1}}}

\newtheorem{definition}{Definition}

\usepackage{xspace}
\makeatletter
\DeclareRobustCommand\onedot{\futurelet\@let@token\@onedot}
\def\@onedot{\ifx\@let@token.\else.\null\fi\xspace}
\def\eg{\emph{e.g}\onedot} 
\def\ie{\emph{i.e}\onedot} 
\def\cf{\emph{c.f}\onedot} 
 
 \def\dof{d.o.f\onedot}
\def\etal{\emph{et al}\onedot}
\makeatother

\makeatletter
\newsavebox\myboxA
\newsavebox\myboxB
\newlength\mylenA
\newcommand*\xoverline[2][0.75]{%
    \sbox{\myboxA}{$\m@th#2$}%
    \setbox\myboxB\null
    \ht\myboxB=\ht\myboxA%
    \dp\myboxB=\dp\myboxA%
    \wd\myboxB=#1\wd\myboxA
    \sbox\myboxB{$\m@th\overline{\copy\myboxB}$}
    \setlength\mylenA{\the\wd\myboxA}
    \addtolength\mylenA{-\the\wd\myboxB}%
    \ifdim\wd\myboxB<\wd\myboxA%
       \rlap{\hskip 0.5\mylenA\usebox\myboxB}{\usebox\myboxA}%
    \else
        \hskip -0.5\mylenA\rlap{\usebox\myboxA}{\hskip 0.5\mylenA\usebox\myboxB}%
    \fi}
\makeatother

\usepackage[pagebackref=true,breaklinks=true,letterpaper=true,colorlinks,bookmarks=false]{hyperref}

\threedvfinalcopy 

\def\threedvPaperID{69} 

\ifthreedvfinal\fi
\begin{document}

\title{Large Scale SfM with the Distributed Camera Model}

\author{Chris Sweeney\thanks{This work was completed while the author was at the University of California, Santa Barbara}\\
University of Washington\\
{\tt\small csweeney@cs.washington.edu}
\and
Victor Fragoso\\
West Virginia University\\
{\tt\small victor.fragoso@mail.wvu.edu}
\and
Tobias H\"ollerer \\
University of California Santa Barbara\\
{\tt\small holl@cs.ucsb.edu}
\and 
Matthew Turk\\
University of California Santa Barbara\\
{\tt\small mturk@cs.ucsb.edu}
}

\maketitle

\begin{abstract}

We introduce the distributed camera model, a novel model for Structure-from-Motion (SfM). This model describes image observations in terms of light rays with ray origins and directions rather than pixels. As such, the proposed model is capable of describing a single camera or multiple cameras simultaneously as the collection of all light rays observed. We show how the distributed camera model is a generalization of the standard camera model and we describe a general formulation and solution to the absolute camera pose problem that works for standard or distributed cameras. The proposed method computes a solution that is up to 8 times more efficient and robust to rotation singularities in comparison with gDLS\cite{sweeney2014gdls}. Finally, this method is used in an novel large-scale incremental SfM pipeline where distributed cameras are accurately and robustly merged together. This pipeline is a direct generalization of traditional incremental SfM; however, instead of incrementally adding one camera at a time to grow the reconstruction the reconstruction is grown by adding a \textit{distributed camera}. Our pipeline produces highly accurate reconstructions efficiently by avoiding the need for many bundle adjustment iterations and is capable of computing a 3D model of Rome from over 15,000 images in just 22 minutes.

\end{abstract}

\section{Introduction}

The problem of determining camera position and orientation given a set of correspondences between image observations and known 3D points is a fundamental problem in computer vision. This set of problems has a wide range of applications in computer vision, including camera calibration, object tracking, simultaneous localization and mapping (SLAM), augmented reality, and structure-from-motion (SfM). Incremental SfM is commonly used to create a 3D model from a set of images by sequentially adding images to an initial coarse model thereby ``growing" the model incrementally~\cite{snavely2006photo}. This incremental process is extremely robust at computing a high quality 3D model due to many opportunities to remove outliers via RANSAC and repeated use of bundle adjustment to reduce errors from noise. 

\begin{figure}[t]
\centering
\includegraphics[width=0.8\linewidth,keepaspectratio]{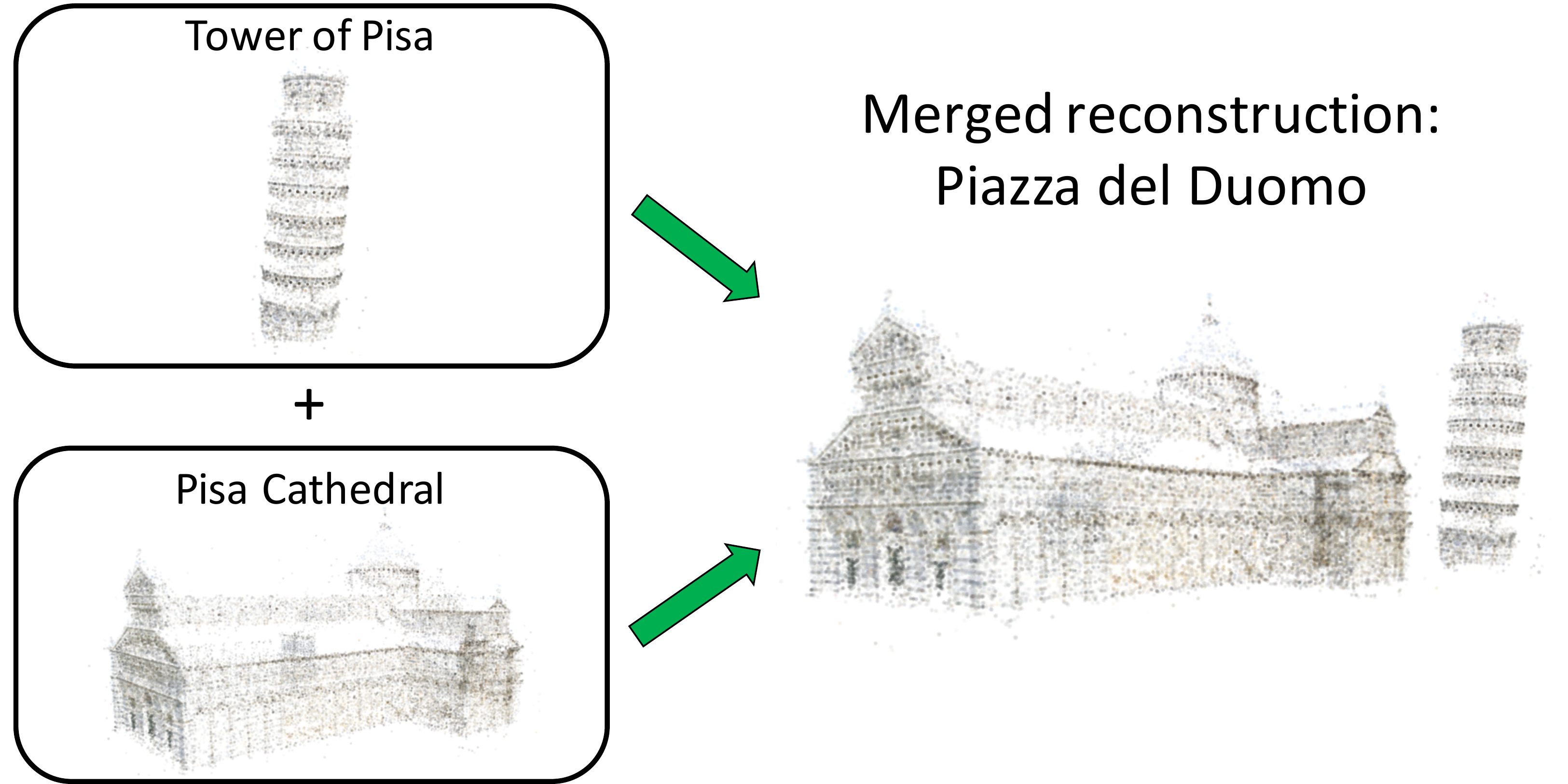}
\caption{ \label{fig:merge_reconstructions}
We are able to merge individual reconstructions by representing multiple cameras as a single distributed camera. The proposed merging process localizes the distributed camera to the current 3D model by solving the generalized absolute pose-and-scale problem.
}
\vspace{-0.1in}
\end{figure}

A core limitation of incremental SfM is its poor scalability. At each step in incremental SfM, the number of cameras in a model is increased by  $\Delta$. For standard incremental SfM pipelines $\Delta = 1$ because only one camera is added to the model at a time. In this paper, we propose to increase the size of $\Delta$, thereby increasing the rate at which we can grow models, by introducing a novel camera parameterization called the \textit{distributed camera model}. The distributed camera model encapsulates image and geometric information from one or multiple cameras by describing pixels as a collection of light rays. As such, observations from multiple cameras may be described as a single distributed camera. 

\begin{definition}
\label{def:distributed_camera}
A \textit{distributed camera} is a collection of observed light rays coming from one or more cameras, parameterized by the ray origins $c_i$, directions $\xoverline{x_i}$, and a single scale parameter for the distributed camera.
\end{definition}

The distributed camera model is similar to the generalized camera model~\cite{pless2003using} with the important distinction that the scale of a distributed camera is unknown and must be recovered\footnote{While a generalized camera~\cite{pless2003using} may not explicitly include scale information, the camera model and accompanying pose methods ~\cite{chen2004pose, kneip2014upnp, nister2007minimal} assume that the scale is known.}. It is also a direct generalization of the standard camera model which occurs when all $c_i$ are equal (i.e. all light rays have the same origin). If we can use distributed cameras in incremental SfM, we can effectively increase the size of $\Delta$. This is because we can add multiple cameras (represented as a single distributed camera) in a single step. This dramatically improves the scalability of the incremental SfM pipeline since it grows models at a faster rate. Even better, as described in Section~\ref{sec:sfm} the proposed SfM pipeline grows $\Delta$ at an exponential rate leading to an extremely efficient and scalable reconstruction method.

In order to use the distributed camera model for incremental SfM, we must determine how to add distributed cameras to the current model. While standard incremental SfM pipelines adds a camera at a time by solving for its absolute pose from 2D-3D correspondences, the proposed method adds a distributed camera by solving the generalized pose-and-scale  problem from 2D-3D correspondences. As part of this work, we show that the generalized pose-and-scale problem is a generalization of the P$n$P problem to multiple cameras which are represented by a distributed camera; we recover the position and orientation as well as the internal scale of the distributed camera with respect to known 3D points. Our solution method improves on previous work~\cite{sweeney2014gdls}  by using the Gr\"obner basis technique to compute the solution more accurately and efficiently. Experiments on synthetic and real data show that our method is more accurate and scalable than other alignment methods. 

We show the applicability of the distributed camera model for incremental SfM in a novel incremental pipeline that utilizes the generalized pose-and-scale method for model-merging. We call our method the generalized Direct Least Squares+++ (gDLS+++) solution, as it is a generalization of the DLS PnP algorithm presented in~\cite{hesch2011direct} and an extension of the gDLS algorithm~\cite{sweeney2014gdls}\footnote{The ``+++" is commonly used in literature to denote a non-degenerate rotation parameterization~\cite{hesch2011direct}}. Our pipeline achieves state-of-the-art results on large scale datasets and reduces the need for expensive bundle adjustment iterations.

In this paper, we make the following contributions:
\begin{enumerate}
\item A new camera model for 3D reconstruction: the \textit{distributed camera model}.
\item Theoretical insight to the absolute pose problem. We show that the gPnP+s problem~\cite{sweeney2014gdls} is in fact a generalization of the standard absolute pose problem and show that its solution is capable of solving the absolute pose problem.\texttt{•}
\item An improved solution method compared to ~\cite{sweeney2014gdls}. By using techniques from UPnP~\cite{kneip2014upnp}, we are able to achieve a much faster and more efficient solver than~\cite{sweeney2014gdls}.
\item A novel incremental SfM pipeline that uses the distributed camera model to achieve large improvements in efficiency and scalability. This pipeline is capable of reconstructing Rome from over 15,000 images in just 22 minutes on a standard desktop computer.
\end{enumerate}

\section{Related Work}
\label{sec:related_work}

\begin{figure*}[t]
\centering
\includegraphics[width=0.27\linewidth,keepaspectratio]{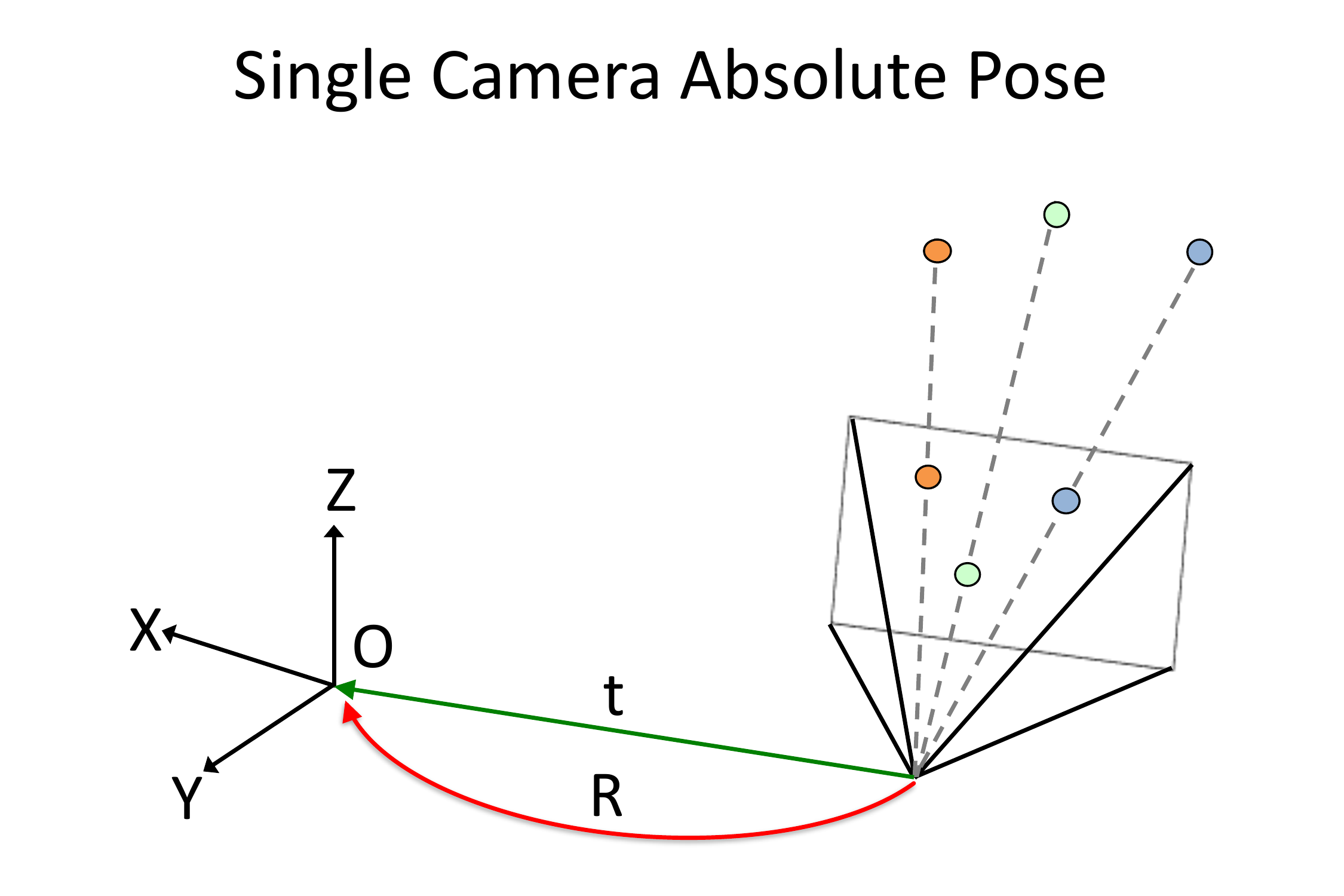}
\includegraphics[width=0.27\linewidth,keepaspectratio]{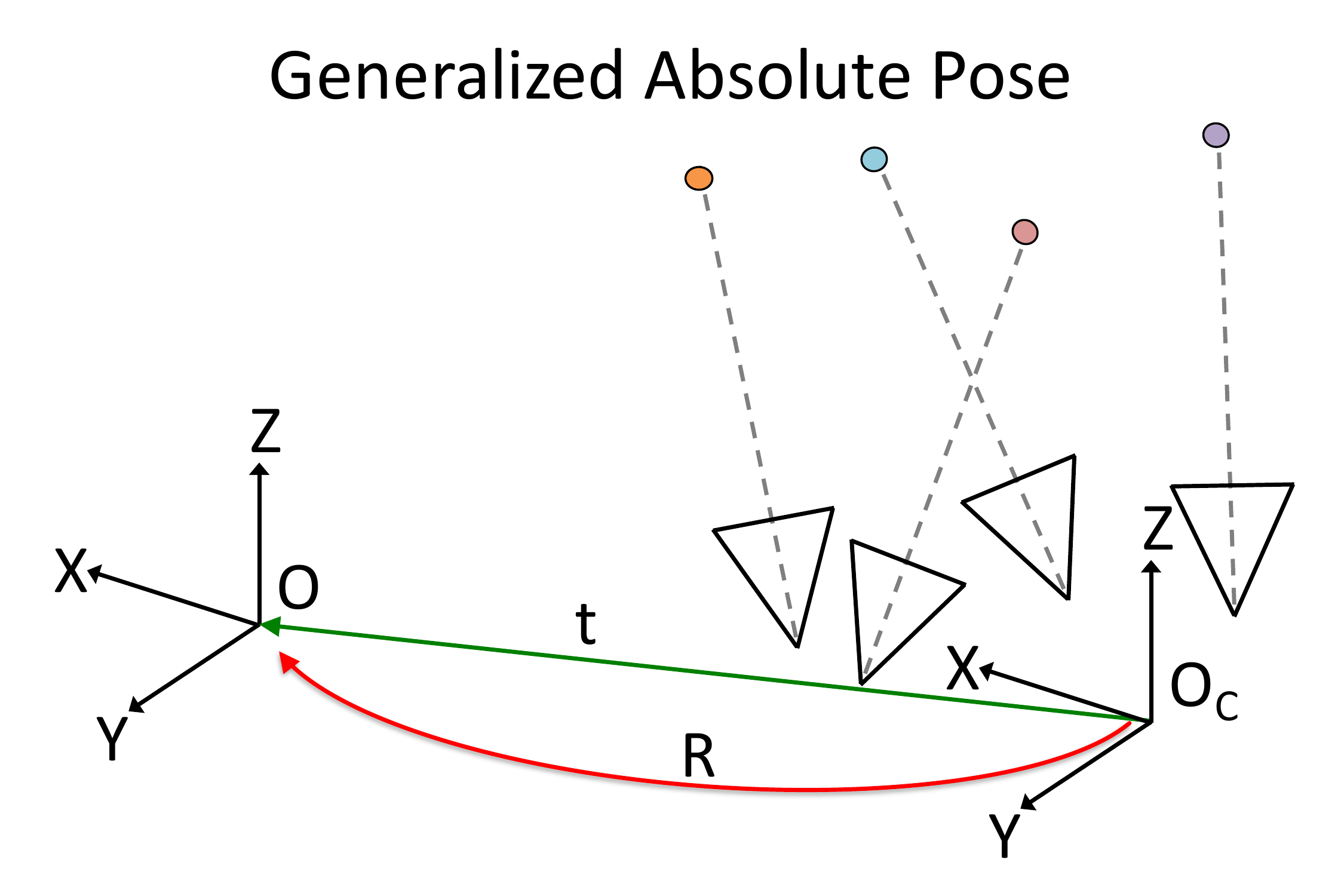}
\includegraphics[width=0.27\linewidth,keepaspectratio]{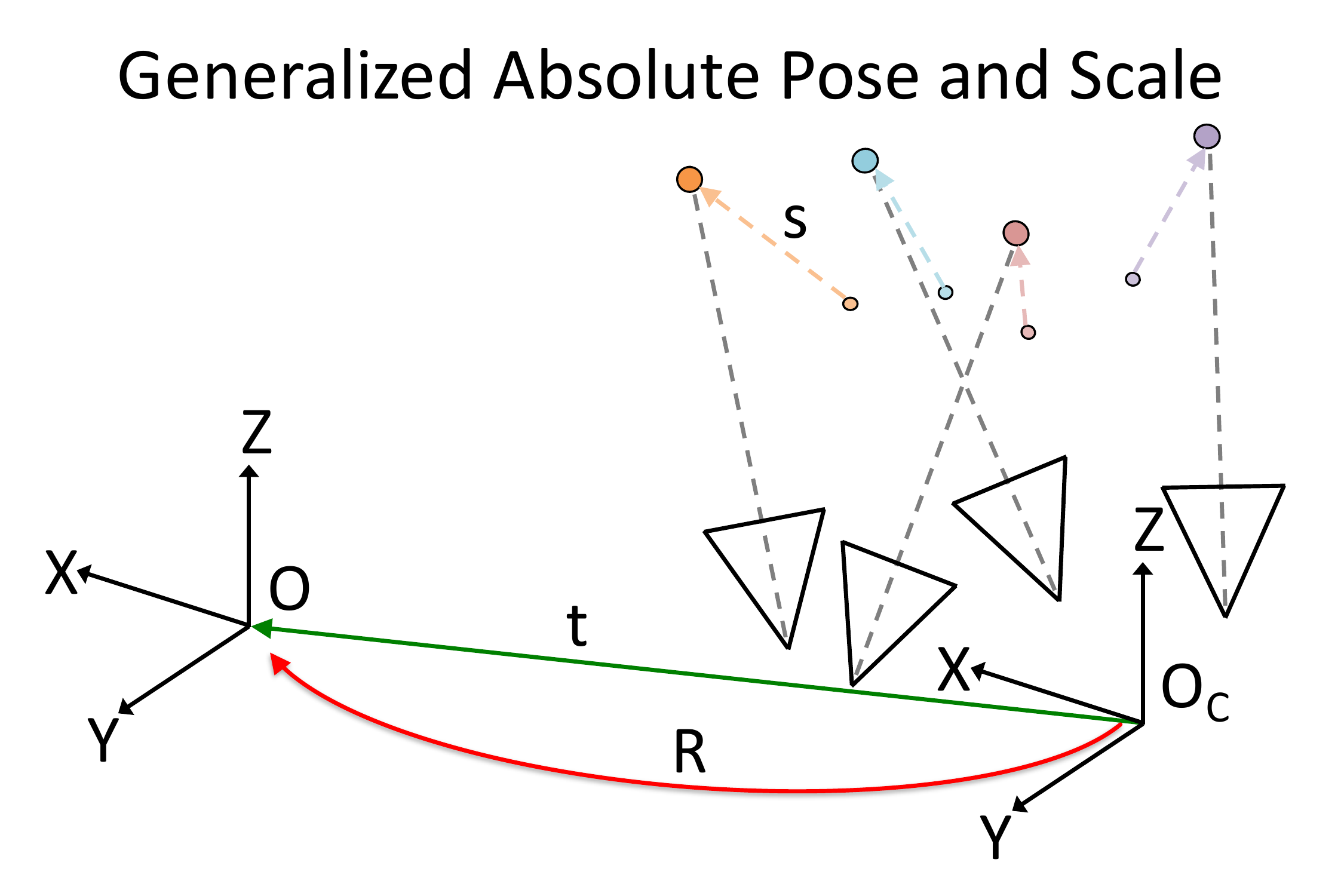}
\caption{ \label{fig:absolute_pose}
The absolute camera pose problem determines a camera's position and orientation with respect to a coordinate system with an origin $O$ from correspondences between 2D image points and 3D points. In this paper, we show how solving the generalized absolute pose-and-scale (right) is a direct generalization of solving the single-camera (left) and multi-camera absolute pose problems.}
\vspace{-0.1in}
\end{figure*}

Since the seminal work of Phototourism~\cite{snavely2006photo}, incremental SfM has been a popular pipeline for 3D reconstruction of unorganized photo collections. Robust open source solutions such as Bundler~\cite{snavely2006photo} and VisualSfM exist~\cite{wu2013towards} using largely similar strategies to grow 3D models by successively adding one image at a time and carefully performing filtering and refinement. These pipelines have limited scalability but display excellent robustness and accuracy. In contrast, global SfM techniques ~\cite{govindu2001combining, jiang2013global, wilson2014robust} are able to compute all camera poses simultaneously, leading to extremely efficient solvers for large-scale problems; however, these methods lack robustness and are typically less accurate than incremental SfM. Alternatively, hierarchical SfM methods compute 3D reconstructions by first breaking up the input images into clusters that are individually reconstructed then merged together into a common 3D coordinate system \cite{divide2014accv}. Typically, bundle adjustment is run each time clusters are merged. This optimization ensures good accuracy and robustness but still a more scalable overall pipeline since fewer instances of bundle adjustment are performed compared to incremental SfM.

Computing camera pose is a fundamental step in 3D reconstruction. 
There is much recent work on solving for the camera pose of calibrated cameras~\cite{kneip2011novel, bujnak2008general, fischler1981random, kukelova2008automatic, kukelova2012polynomial}. Extending the single-camera absolute pose problem to multi-camera rigs is the Non-Perspective-$n$-Point (NP$n$P) problem. Minimal solutions to the NP3P problem were proposed by N\'ister and St\'ewenius~\cite{nister2007minimal} and Chen and Chang~\cite{chen2004pose} for generalized cameras. The UP$n$P algorithm~\cite{kneip2014upnp} is an extension of the DLS P$n$P algorithm~\cite{hesch2011direct} that is capable of solving the single-camera or multi-camera absolute pose problem; however, it does not estimate scale and therefore cannot be used with distributed cameras.

To additionally recover scale transformations in multi-camera rigs, Ventura \etal~\cite{ventura2014cvpr} presented the first minimal solution to the generalized pose-and-scale problem. They use the generalized camera model and employ Gr\"obner basis computations to efficiently solve for scale, rotation, and translation using 4 2D-3D correspondences. Sweeney \etal~\cite{sweeney2014gdls} presented the first nonminimal solver, gDLS, for the generalized pose and scale problem. By extending the Direct Least Squares (DLS) P$n$P algorithm of Hesch and Roumeliotis~\cite{hesch2011direct}, rotation, translation, and scale can be solved for in a fixed size polynomial system. While this method is very scalable the large elimination template of the gDLS method makes it very inefficient and the Cayley rotation parameterization results in singularities. In this paper, we extend the gDLS solution method to a non-singular representation and show that using the Gr\"obner basis technique achieves an 8$\times$ speedup.

\section{The Absolute Camera Pose Problem}
\label{sec:absolute_pose}

In this section we review the absolute camera pose problem and demonstrate how to generalize the standard formulation to distributed cameras.

The fundamental problem of determining camera position and orientation given a set of  correspondences between 2D image observations and known 3D points is called the \textit{absolute camera pose problem} or Perspective n-Point (PnP) problem (\cf Figure~\ref{fig:absolute_pose} left). In the minimal case, only three 2D-3D correspondences are required to compute the absolute camera pose~\cite{fischler1981random, hesch2011direct, kneip2011novel}. These methods utilize the reprojection constraint such that 3D points $X_i$ align with unit-norm pixel rays $\xoverline{x_i}$ when rotated and translated:

\begin{equation}
\label{eq:reprojection_error}
\alpha_i \xoverline{x_i} = R X_i + t, \qquad i=1,2,3
\end{equation}
where $R$ and $t$ rotate and translate 3D points into the camera coordinate system. The scalar $\alpha_i$ stretches the unit-norm ray $x_i$ such that $\alpha_i = ||R X_i +  t||$. In order to determine the camera's pose, we would like to solve for the unknown rotation $R$ and translation $t$ that minimize the reprojection error:
\begin{equation}
\label{eq:single_camera_cost_function}
C(R, t) = \sum_{i=1}^{3}||\xoverline{x_i} - \dfrac{1}{\alpha_i}(R X_i + t)||^2
\end{equation}

The cost function $C(R, t)$ is the sum of squared reprojection errors and is the basis for solving the absolute camera pose problem.

\subsection{Generalization to 7 \dof}

When information from multiple images is available, P$n$P methods are no longer suitable and few methods exist that are able to jointly utilize information from many cameras simultaneously. As illustrated in Figure \ref{fig:absolute_pose} (center), multiple cameras (or multiple images from a single moving camera) can be described with the generalized camera model~\cite{pless2003using}. The generalized camera model represents a set of observations by the viewing ray origins and directions. For multi-camera systems, these values may be determined from the positions and orientations of cameras within the rig. The generalized camera model considers the viewing rays as static with respect to a common coordinate system (\cf $O_C$ in Figure ~\ref{fig:absolute_pose} center, right). Using the generalized camera model, we may extend the reprojection constraint of Eq.~\eqref{eq:reprojection_error} to multiple cameras:
\begin{equation}
c_i + \alpha_i \xoverline{x_i} = R X_i + t, \qquad i=1,\ldots,n
\end{equation}
where $c_i$ is the origin of the feature ray $\xoverline{x_i}$ within the generalized camera model. This representation assumes that the scale of the generalized camera is equal to the scale of the 3D points (\eg, that both have metric scale). In general, the internal scale of each generalized camera is not guaranteed to be consistent with the scale of the 3D points. Consider a multi-camera system on a car that we want to localize to a point cloud created from Google Street View images. While we may know the metric scale of the car's multi-camera rig, it is unlikely we have accurate scale calibration for the Google Street View point cloud, and so we must recover the scale transformation between the rig and the point cloud in addition to the rotation and translation. When the scale is unknown then we have a distributed camera (\cf Definition~\ref{def:distributed_camera}). This leads to the following reprojection constraint for distributed cameras that accounts for scale:
\begin{equation} \label{eq:reprojection_constraint}
s c_i + \alpha_i \xoverline{x_i} = R X_i + t, \qquad i=1,\ldots,n
\end{equation}
where $\alpha_i$ is a scalar which stretches the image ray such that it meets the world point $X_i$ such that $\alpha_i = ||R X_i + t -s c_i||$. Clearly the generalized absolute pose problem occurs when $s =1$ and the single-camera absolute pose problem occurs when $c_i = \textbf{0} \;\; \forall i$.

By extending Eq~\eqref{eq:reprojection_error} to multi-camera systems and scale transformations, we have generalized the PnP problem to the generalized pose-and-scale (gPnP+s) problem in Eq~\eqref{eq:reprojection_constraint} shown in Figure~\ref{fig:absolute_pose} (right). The goal of the gPnP+s problem is to determine the pose and internal scale of a distributed camera with respect to $n$ known 3D points. This is equivalent to aligning the two coordinate systems that define the distributed camera and the 3D points. Thus the solution to the gPnP+s problem is a 7 \dof similarity transformation. 

\section{An $L_2$ Optimal Solution}
\label{sec:solution_method}

To solve the generalized pose-and-scale problem we build on the method of Sweeney \etal~\cite{sweeney2014gdls}, making modifications to the solver based on the UPnP method~\cite{kneip2014upnp}. We extend the method of ~\cite{sweeney2014gdls} with the following ideas from UPnP~\cite{kneip2014upnp} to achieve a faster and more accurate solver:
\begin{enumerate}
\item Using the quaternion representation for rotations. This avoids the singularity of the Cayley-Gibbs-Rodrigues parameterization and improves accuracy.
\item Solve the system of polynomials using the Gr\"obner basis technique instead of the Macaulay Matrix method.
\item Take advantage of p-fold symmetry to reduce the size of the polynomial system~\cite{ask2012pfold}.
\end{enumerate}

We briefly review the solution method. When considering all $n$ 2D-3D correspondences, there exists $8 + n$ unknown variables (4 for quaternion rotation, 3 for translation, 1 for scale, and 1 unknown depth per observation). The gP$n$P+s problem can be formulated from Eq.~\eqref{eq:reprojection_constraint} as a non-linear least-squares minimization of the measurement errors. Thus, we aim to minimize the cost function:

\begin{equation}
\label{eq:cost_function}
C(R,t,s) = \sum_{i=1}^n ||\xoverline{x_i} - \frac{1}{\alpha_i}(R X_i + t - s c_i)||^2.
\end{equation}

The cost function shown in Eq.~\eqref{eq:cost_function} can be minimized by a least-square solver. However, we can rewrite the problem in terms of fewer unknowns. Specifically, we can rewrite this equation solely in terms of the unknown rotation, $R$. When we relax the constraint that  $\alpha_i = ||R X_i + t -s c_i||$ and treat each $\alpha_i$ as a free variable, $\alpha_i$, $s$, and $t$ appear linearly and can be easily reduced from Eq.~\eqref{eq:cost_function}. This relaxation is reasonable since solving the optimality conditions results in $\alpha_i^* = z_i^\top(R X_i + t -s c_i)$ where $z_i$ is $\xoverline{x_i}$ corrupted by measurement noise.

We begin by rewriting our system of equations from Eq.~\eqref{eq:reprojection_constraint} in matrix-vector form:
\begin{align}\label{eq:matrix_block_constraint}
\underbrace{
\begin{bmatrix}
\xoverline{x_1} & \; & \; & c_1 & -I \\
\; & \ddots & \; & \vdots & \vdots \\
\; & \; & \xoverline{x_n} & c_n & -I
\end{bmatrix}}_{\text{A}}
\underbrace{
\begin{bmatrix}
\alpha_1 \\
\vdots \\
\alpha_n \\
s \\
t
\end{bmatrix}}_{\text{x}}
&= 
\underbrace{
\begin{bmatrix}
R & \; & \;  \\
\; & \ddots & \;  \\
\; & \; & R 
\end{bmatrix}}_{\text{W}}
\underbrace{
\begin{bmatrix}
X_1 \\
\vdots \\
X_n
\end{bmatrix}}_{\text{b}} \\
\Leftrightarrow Ax &= Wb,
\end{align}
where $A$ and $b$  consist of known and observed values, $x$ is the vector of unknown variables we will eliminate from the system of equations, and $W$ is the block-diagonal matrix of the unknown rotation matrix. From Eq.~\eqref{eq:matrix_block_constraint}, we can create a simple expression for x:
\begin{equation}
x = (A^\top A)^{-1}A^\top Wb = \begin{bmatrix} U \\ S \\ V \end{bmatrix} Wb.
\end{equation}
We have partitioned $(A^\top A)^{-1}A^\top$ into constant matrices $U$, $S$, and $V$ such that the depth, scale, and translation parameters are functions of $U$, $S$, and $V$ respectively. Matrices $U$, $S$, and $V$ can be efficiently computed in closed form by exploiting the sparse structure of the block matrices (see Appendix A from ~\cite{sweeney2014gdls} for the full derivation). Note that $\alpha_i$, $s$, and $t$ may now be written concisely as linear functions of the rotation:
\begin{align}
\label{eq:st_substitution}
\alpha_i &= u_i^\top W b \\
s &= SWb \\
t &= VWb,
\end{align}
where $u_i^\top$ is the $i$-th row of $U$. Through substitution, the geometric constraint equation~\eqref{eq:reprojection_constraint} can be rewritten as:
\begin{equation} \label{eq:linear_geometric_constraint}
\underbrace{SWb}_s c_i + \underbrace{u_i^\top Wb}_{\alpha_i} \xoverline{x_i} = RX_i + \underbrace{VWb}_t.
\end{equation}
This new constraint is quadratic in the four rotation unknowns given by the unit-norm quaternion representation.
\subsection{A Least Squares Cost Function}
\label{sec:new_cost_function}
The geometric constraint equation~\eqref{eq:linear_geometric_constraint} assumes noise-free observations. We assume noisy observations $\xoverline{z_i} = \xoverline{x_i} + \eta_i$, where $\eta_i$ is zero mean noise. Eq.~\eqref{eq:linear_geometric_constraint} may be rewritten in terms of our noisy observation:
\begin{align}
SWbc_i + u_i^\top Wb (\xoverline{z_i} - \eta_i) = RX_i + VWb \\
\label{eq:noisy_ls_term}
\Rightarrow \eta_i' = SWbc_i + u_i^\top Wb\xoverline{z_i} - RX_i - VWb,
\end{align}
where $\eta_i'$ is a zero-mean noise term that is a function of $\eta_i$ (but whose covariance depends on the system parameters, as noted by Hesch and Roumeliotis~\cite{hesch2011direct}). We evaluate $u_i$, $S$, and $V$ at $\xoverline{x_i}=\xoverline{z_i}$ without loss of generality. Observe that $u_i$ can be eliminated from Eq.~\ref{eq:noisy_ls_term} by noting that:
\begin{align}
&UWb = \begin{bmatrix}
\xoverline{z_i}^\top & \; & \; \\
\; & \ddots & \; \\
\; & \; & \xoverline{z_n}^\top
\end{bmatrix}Wb
- \begin{bmatrix}
\xoverline{z_1}^\top c_1 \\ \vdots \\ \overline{z_n}^\top c_n
\end{bmatrix} SWb +
\begin{bmatrix}
\overline{z_1}^\top \\ \vdots \\ \overline{z_n}^\top
\end{bmatrix} VWb \\
\label{eq:expanded_u}
&\Rightarrow u_i^\top Wb = \overline{z_i}^\top R X_i - \overline{z_i}^\top c_i SWbc_i + \overline{z_i}^\top VWb.
\end{align}
 Through substitution, Eq.~\eqref{eq:noisy_ls_term} can be refactored such that:
\begin{equation}
\label{eq:factored_noisy_ls_term}
\eta_i' = (\overline{z_i}\overline{z_i}^\top - I_3)(RX_i - SWbc_i + VWb).
\end{equation}

Eq.~\eqref{eq:factored_noisy_ls_term} allows the gP$n$P+s problem to be formulated as an unconstrained least-squares minimization in 4 unknown rotation parameters. We formulate the least squares cost function, $C'$, as the sum of the squared constraint errors from Eq. \eqref{eq:factored_noisy_ls_term}:
\begin{align}
\label{eq:ls_sum}
C'(R) &= \sum_{i=1}^n || (\overline{z_i}\overline{z_i}^\top - I_3)(RX_i - SWbc_i + VWb) ||^2 \\
&= \sum_{i=1}^n \eta_i'^\top \eta_i'.
\end{align}
Thus, the number of unknowns in the system has been reduced from $8 + n$ to $4$. This is an important part of the formulation, as it allows the size of the system to be independent of the number of observations and thus scalable. To enforce a solution with a valid rotation we must additionally enforce that the quaternion is unit-norm: $q^\top q = 1.$

\subsection{Gr\"obner Basis Solution}

An alternative method for solving polynomial systems is the Gr\"obner basis technique~\cite{kukelova2008automatic}. We created a Gr\"obner basis solver with an automatic generator similar to \cite{kukelova2008automatic}\footnote{See ~\cite{kukelova2008automatic} for more details about Gr\"obner basis techniques.} while taking advantage of several additional forms of improvement. Following the solver of Kneip \etal ~\cite{kneip2014upnp}, the size of the Gr\"obner basis is reduced by only choosing random values in $\mathbb{Z}_p$ that correspond to valid configurations for the generalized pose-and-scale problem. Next, double-roots are eliminated by exploiting the 2-fold symmetry technique used in \cite{ask2012pfold, kneip2014upnp, zheng2013iccv}. This technique requires that all polynomials contain only even or only odd-degree monomials. The first order optimality constraints (formed from the partial derivitives of $C'$) contain only uneven monomials; however, the unit-norm quaternion constraint contains even monomials. By modifying the unit-norm quaternion constraint to the squared norm:
\begin{equation}
\label{eq:pfold_unit_norm_constraint}
(q^\top q - 1)^2 = 0
\end{equation}
we obtain equations whose first order derivatives contain only odd monomials. Our final polynomial system is then:
\begin{align}
\dfrac{\partial C'}{\partial q_i} &= 0 \qquad i=0,1,2,3 \\
\label{eq:norm_partial}
(q^\top q - 1)q_i &= 0 \qquad i=0,1,2,3.
\end{align}
These eight third-order polynomials contain only uneven degree monomials, and so we can apply the 2-fold symmetry technique proposed by Ask \etal~\cite{ask2012pfold}. As with the UPnP method~\cite{kneip2014upnp}, applying these techniques to our Gr\"obner basis solver creates a 141 $\times$ 149 elimination template with an action matrix that is 8 $\times$ 8. Both the elimination template and the action matrix are dramatically smaller than with the Macaulay Matrix solution of ~\cite{sweeney2014gdls}, leading to a total execution time of just 151 $\mu$s. 
\section{Results}
\label{sec:experiments}

\subsection{Numerical Stability}
\begin{figure}[t]
\includegraphics[width=0.325\linewidth,keepaspectratio]{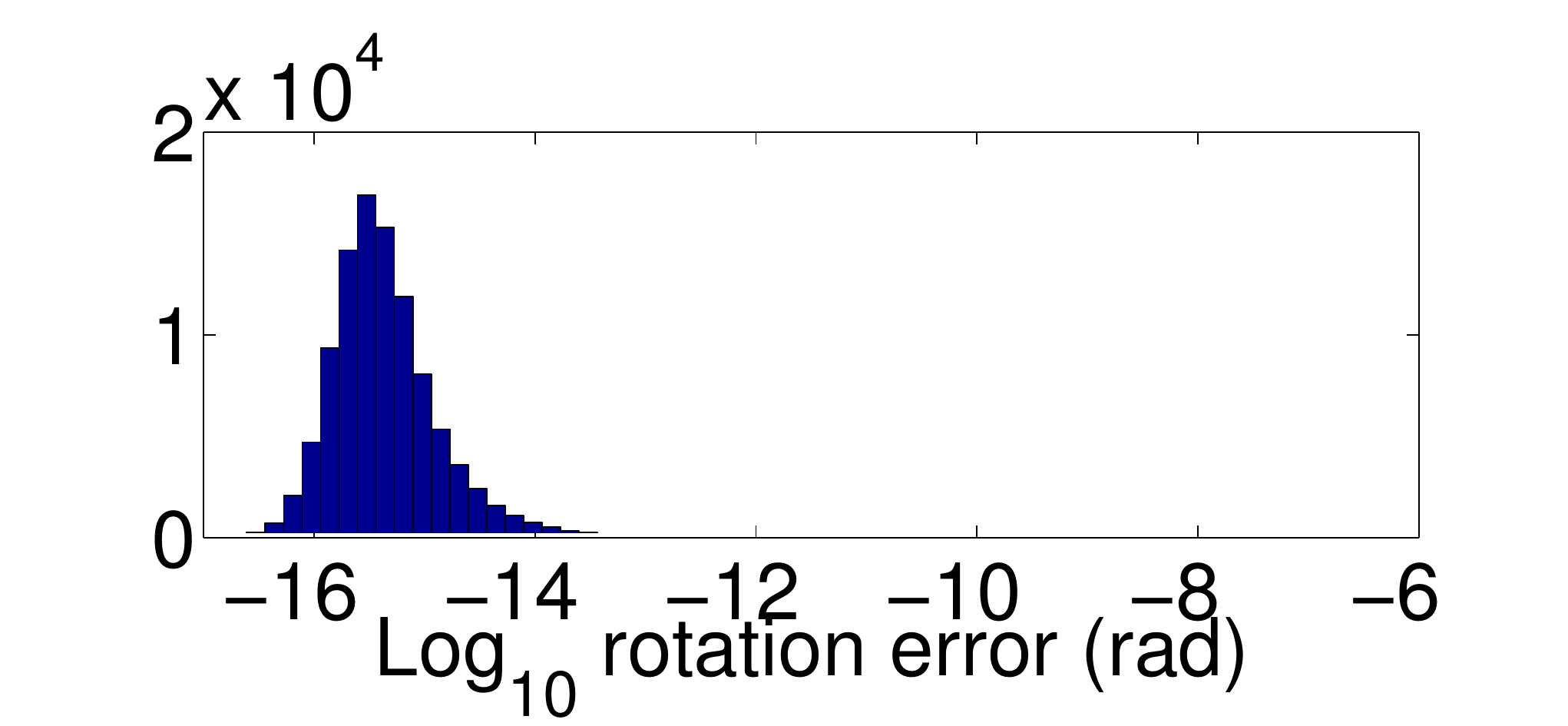}
\includegraphics[width=0.325\linewidth,keepaspectratio]{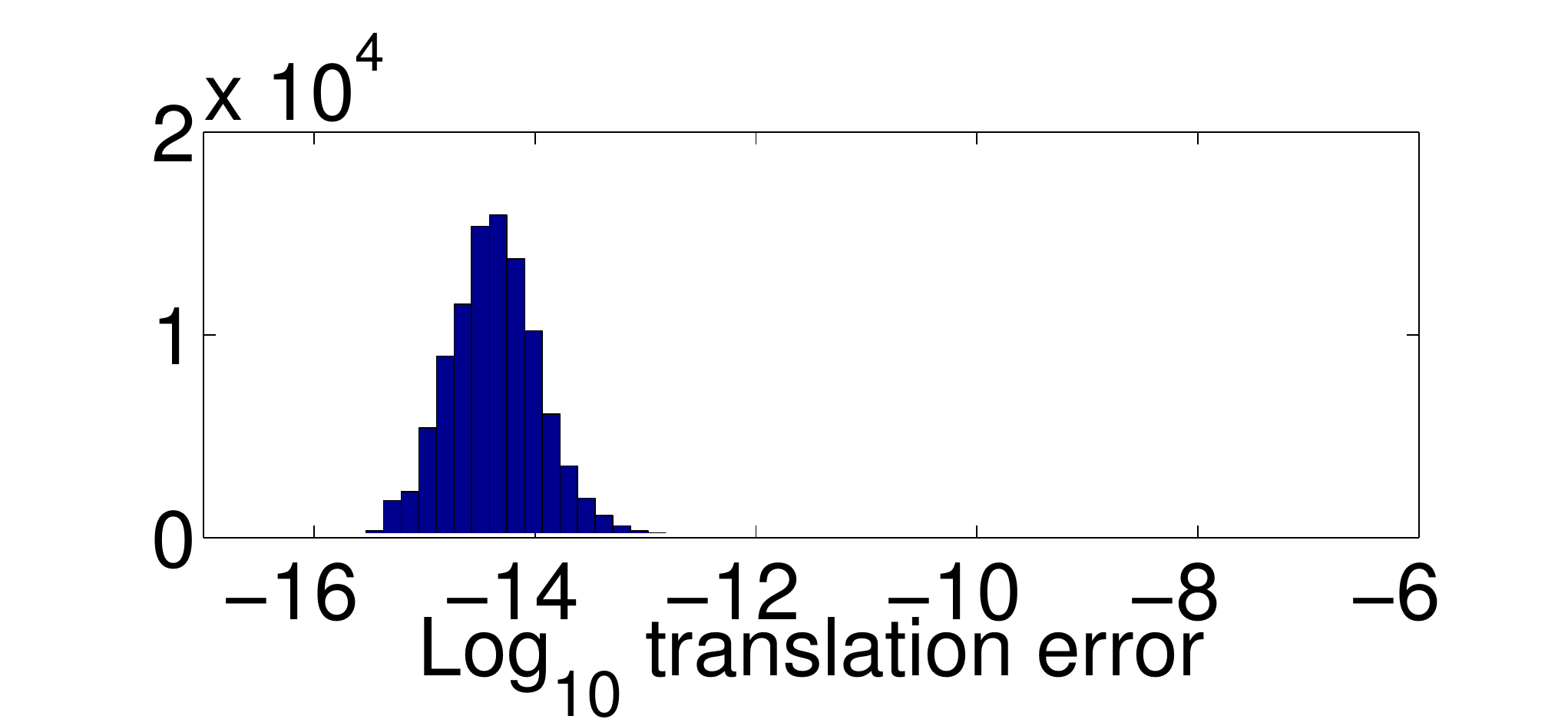}
\includegraphics[width=0.325\linewidth,keepaspectratio]{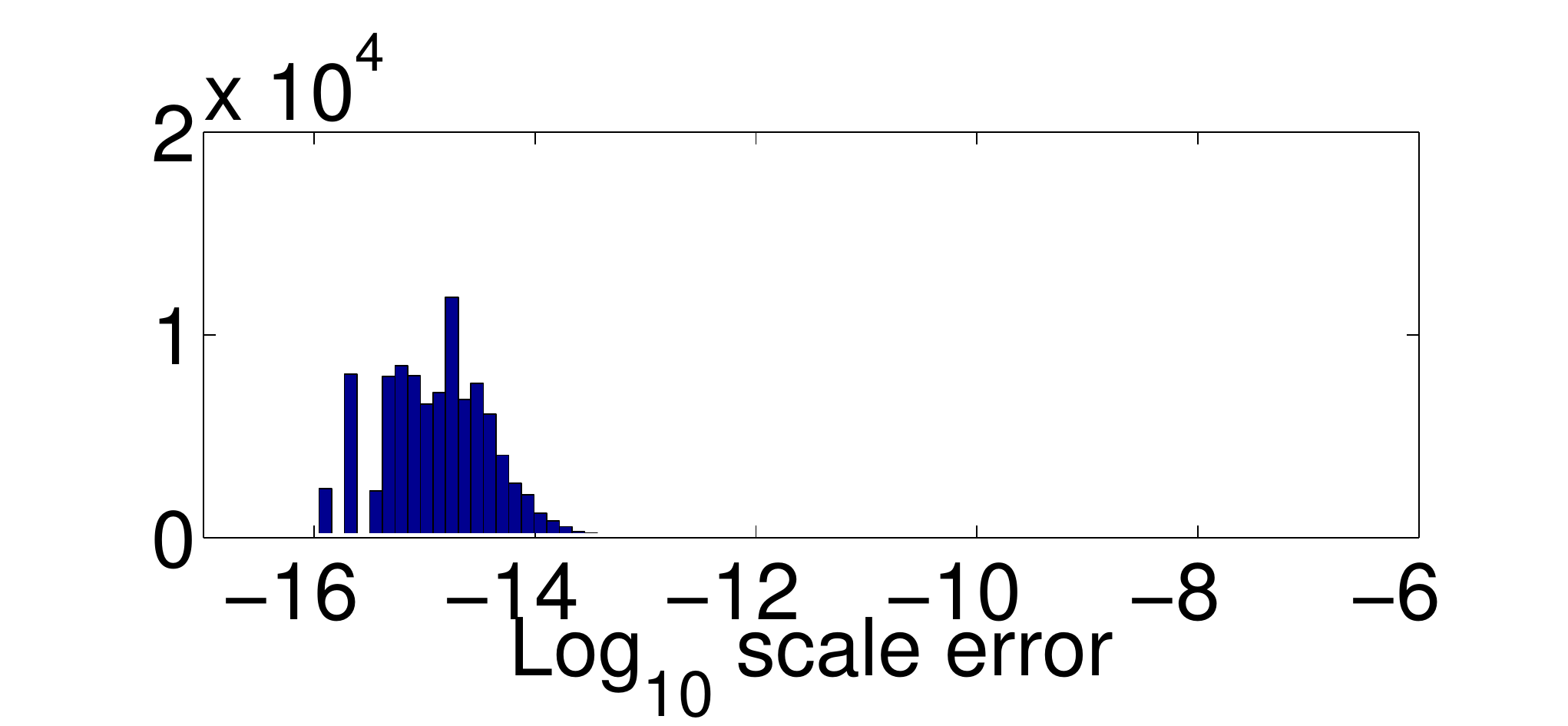} 
\caption{\label{fig:num_stability} 
Histograms of numerical errors in the computed similarity transforms based on $10^5$ random trials with the minimal 4 correspondences. Our algorithm is extremely stable, leading to high accuracy even in the presence of noise.}
\end{figure}

We tested the numerical stability of our solution over $10^5$ random trials. We generated uniformly random camera configurations that placed cameras (\ie, ray origins) in the cube $[-1, 1]\times[-1,1]\times[-1,1]$ around the origin. The 3D points were randomly placed in the volume $[-1,1]\times[-1,1]\times[2,4]$. Ray directions were computed as unit vectors from camera origins to 3D points. An identity similarity transformation was used (\ie, $R=I$, $t=0$, $s=1$). For each trial, we computed solutions using the minimal 4 correspondences. We calculated the angular rotation, translation, and scale errors for each trial, and plot  the results in Figure~\ref{fig:num_stability}. The errors are very stable, with 98\% of all errors less than $10^{-12}$.

\subsection{Simulations With Noisy Synthetic Data}
\label{sec:synthetic_experiments}

\begin{figure*}[t]
\begin{center}
\includegraphics[width=0.27\linewidth,keepaspectratio]{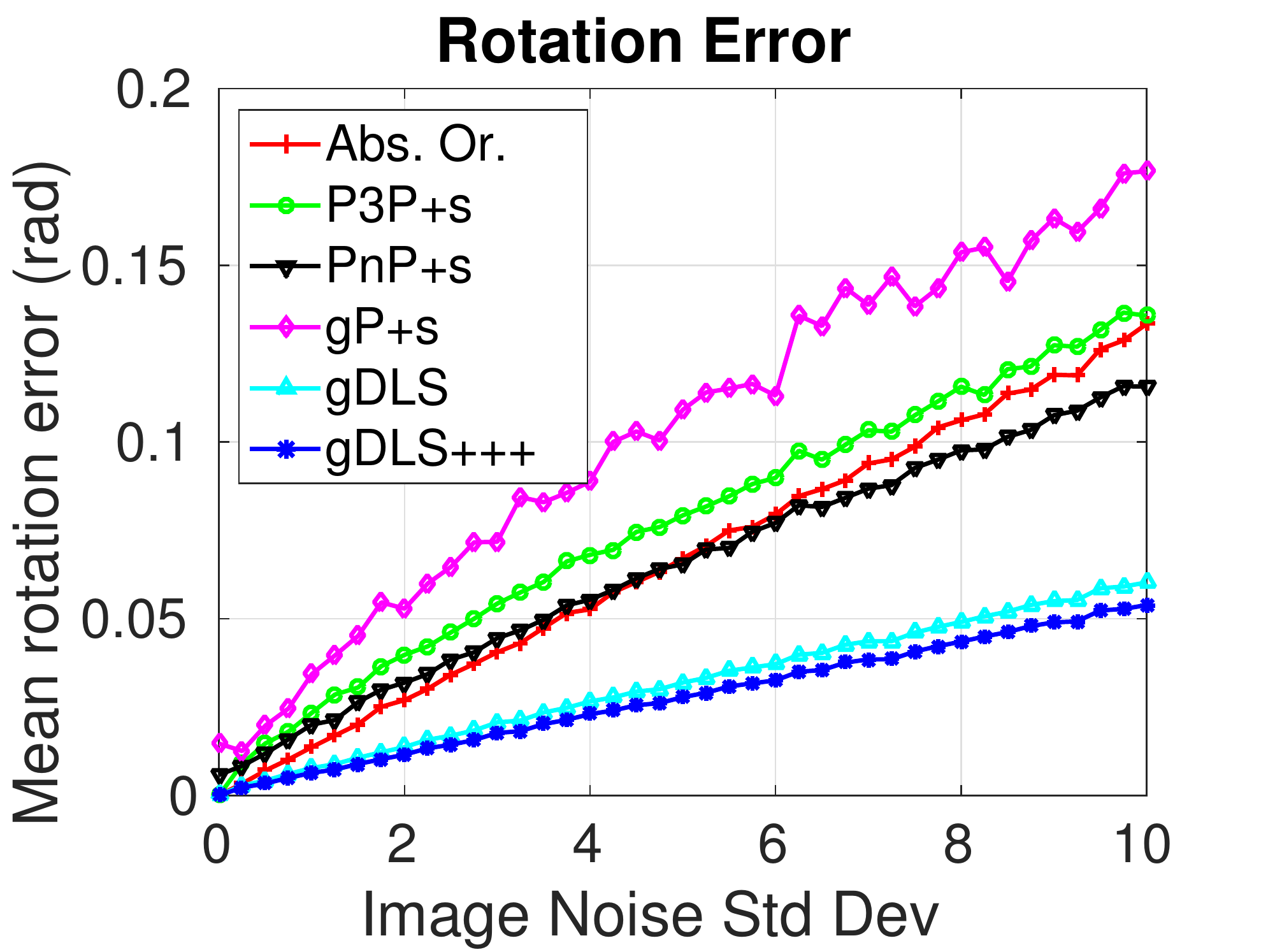}
\includegraphics[width=0.27\linewidth,keepaspectratio]{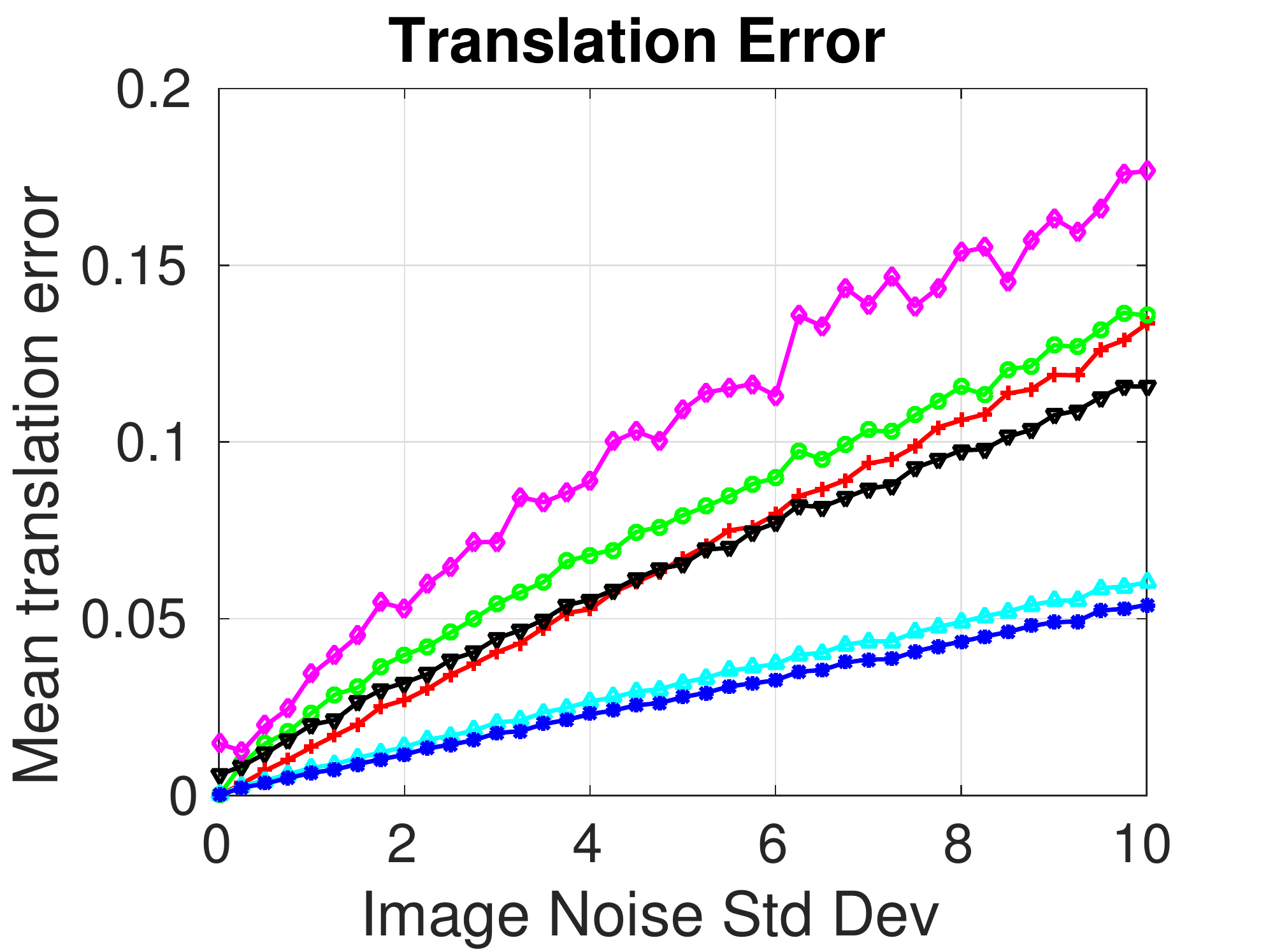}
\includegraphics[width=0.27\linewidth,keepaspectratio]{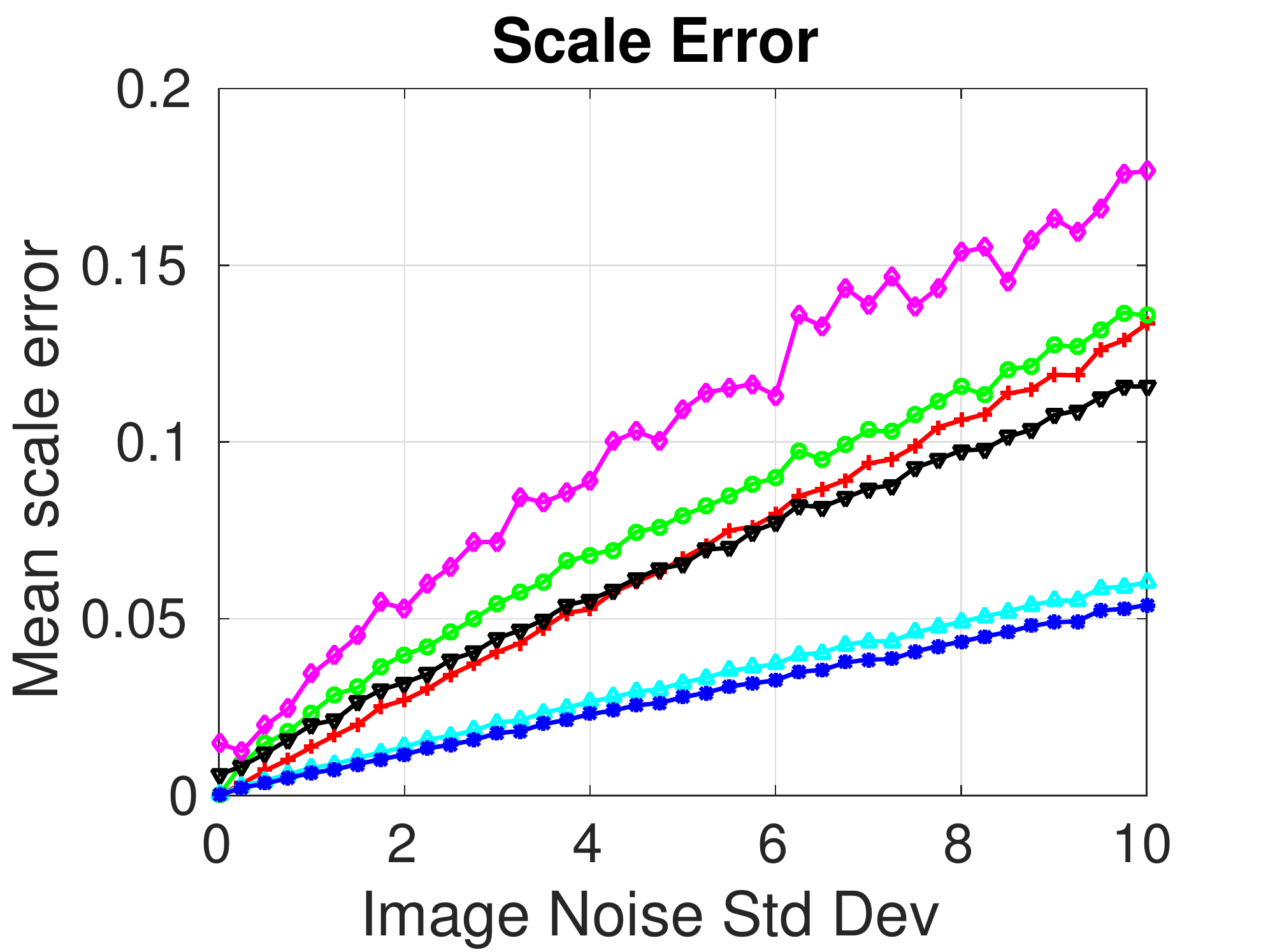}
\end{center}
\vspace{-0.1in}
\caption{\label{fig:image_noise} 
We compared similarity transform algorithms with increasing levels of image noise to measure the pose error performance: the absolute orientation algorithm of Umeyama~\cite{umeyama1991least}, P3P+s, P$n$P+s, gP+s\cite{ventura2014cvpr}, and our algorithm, gDLS. Each algorithm was run with the same camera and point configuration for 1000 trials per noise level. Our algorithm has mean better rotation, translation, and scale errors for all levels of image noise.
\vspace{-0.1in}
}
\end{figure*}

We performed two experiments with synthetic data to analyze the performance of our algorithm as the amount of image noise increases and as the number of correspondences increases. For both experiments we use the same configuration as the numerical stability experiments with six total 2D-3D observations. Using the known 2D-3D correspondences, we apply a similarity transformation with a random rotation in the range of $[-30,30]$ degrees about each of the $x$, $y$, and $z$ axes, a random translation with a distance between $0.5$ and $10$, and a random scale change between $0.1$ and $10$. We measure the performance of the following similarity transform algorithms:
\begin{itemize}
\item \textbf{Absolute Orientation:} The absolute orientation method of Umeyama~\cite{umeyama1991least} is used to align the known 3D points to 3D points triangulated from 2D correspondences. This algorithm is only an alignment method and does not utilize any 2D correspondences.

\item \textbf{P3P+s, P$n$P+s:} First, the scale is estimated from the median scale of triangulated points in each set of cameras. Then, P3P \etal~\cite{kneip2011novel} or P$n$P~\cite{hesch2011direct} is used to determine the rotation and translation. This process is repeated for all cameras, and the camera localization and scale estimation that yields the largest number of inliers is used as the similarity transformation.

\item \textbf{gP+s:} The minimal solver of Ventura \etal~\cite{ventura2014cvpr} is used with 2D-3D correspondences from all cameras. While the algorithm is intended for the minimal case of $n=4$ correspondences, it can compute an overdetermined solution for $n \geq 4$ correspondences.
\item \textbf{gDLS:} The algorithm presented in ~\cite{sweeney2014gdls}, which uses $n\geq 4$ 2D-3D correspondences from all cameras.
\item \textbf{gDLS+++:} The algorithm presented in this paper, which is an extension of the gDLS algorithm~\cite{sweeney2014gdls}. This method uses $n\geq 4$ 2D-3D matches from all cameras.
\end{itemize}

After running each algorithm on the same testbed, we calculate the rotation, translation, and scale errors with respect to the known similarity transformation.

\textbf{Image noise experiment:} We evaluated the similarity transformation algorithms under increased levels of image noise. Using the configuration described above, we increased the image noise from 0 to 10 pixels standard deviation, and ran $1000$ trials at each level. Our algorithm outperforms each of the other similarity transformation algorithms for all levels of image noise, as shown in Figure~\ref{fig:image_noise}. The fact that our algorithm returns all minima of our modified cost function is advantageous under high levels of noise, as we are not susceptible to getting stuck in a bad local minimum. This allows our algorithm to be very robust to image noise as compared to other algorithms.

\begin{figure}[t]
\begin{center}
\includegraphics[width=0.3\linewidth,keepaspectratio]{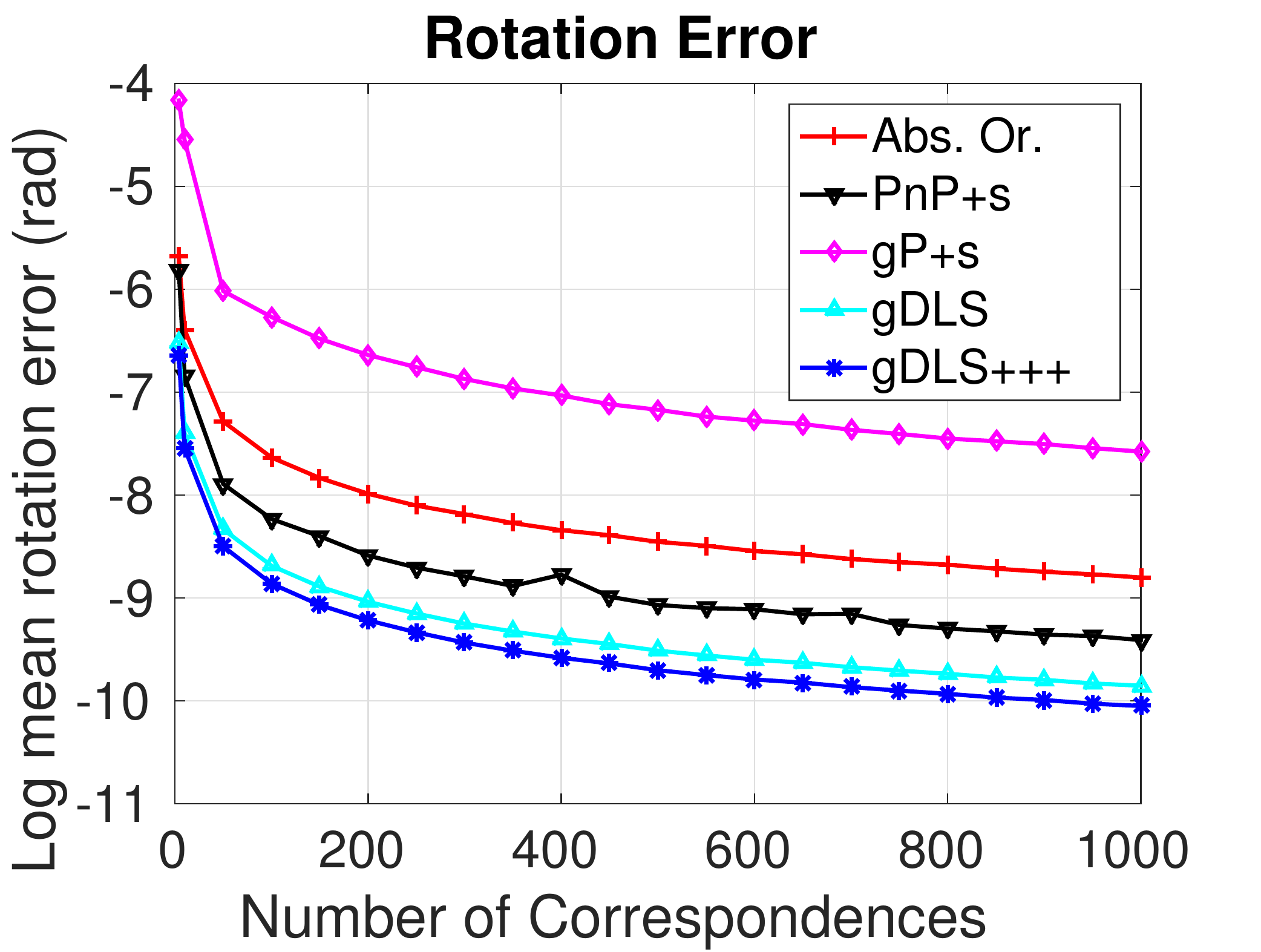}
\includegraphics[width=0.3\linewidth,keepaspectratio]{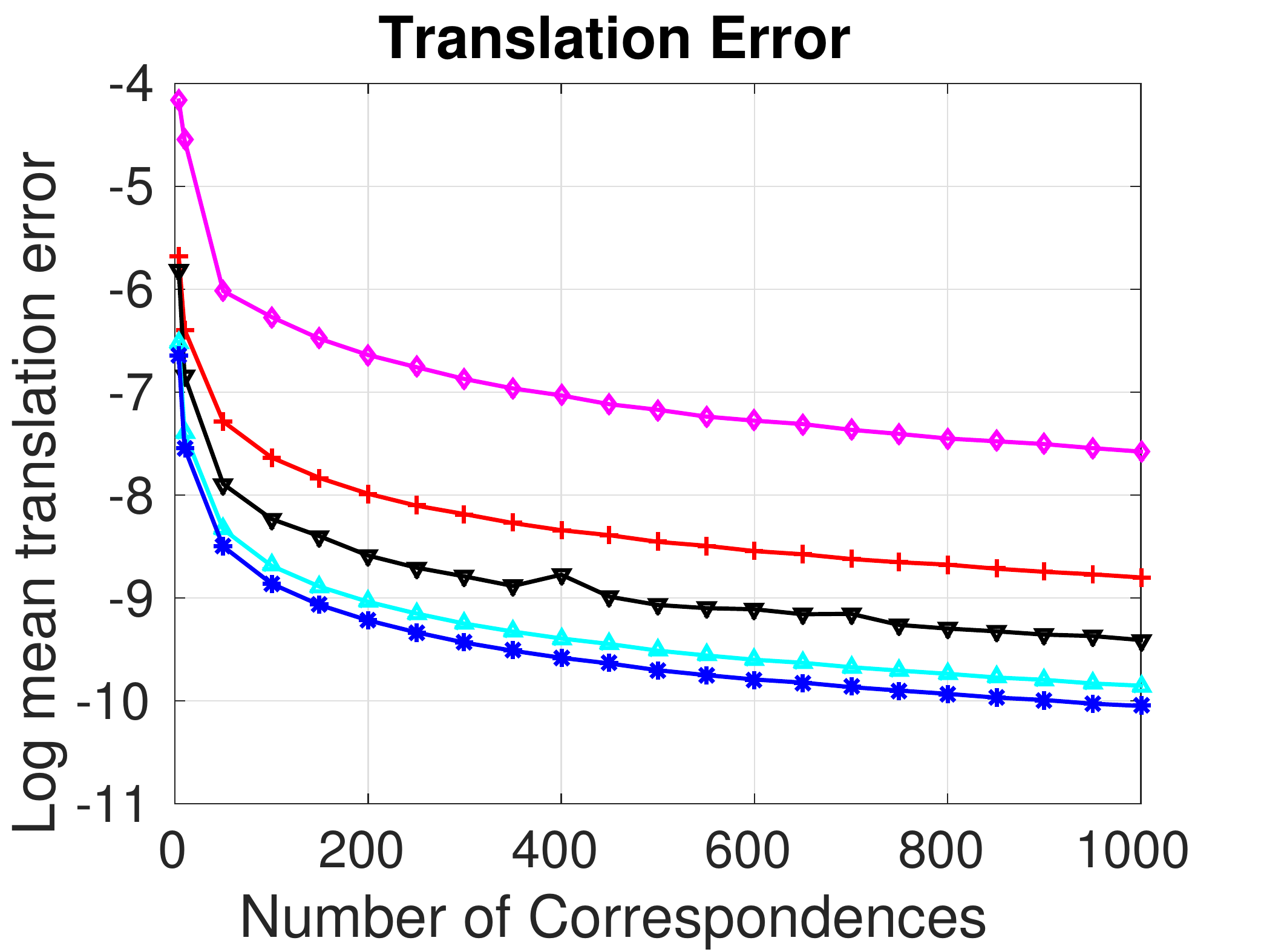}
\includegraphics[width=0.3\linewidth,keepaspectratio]{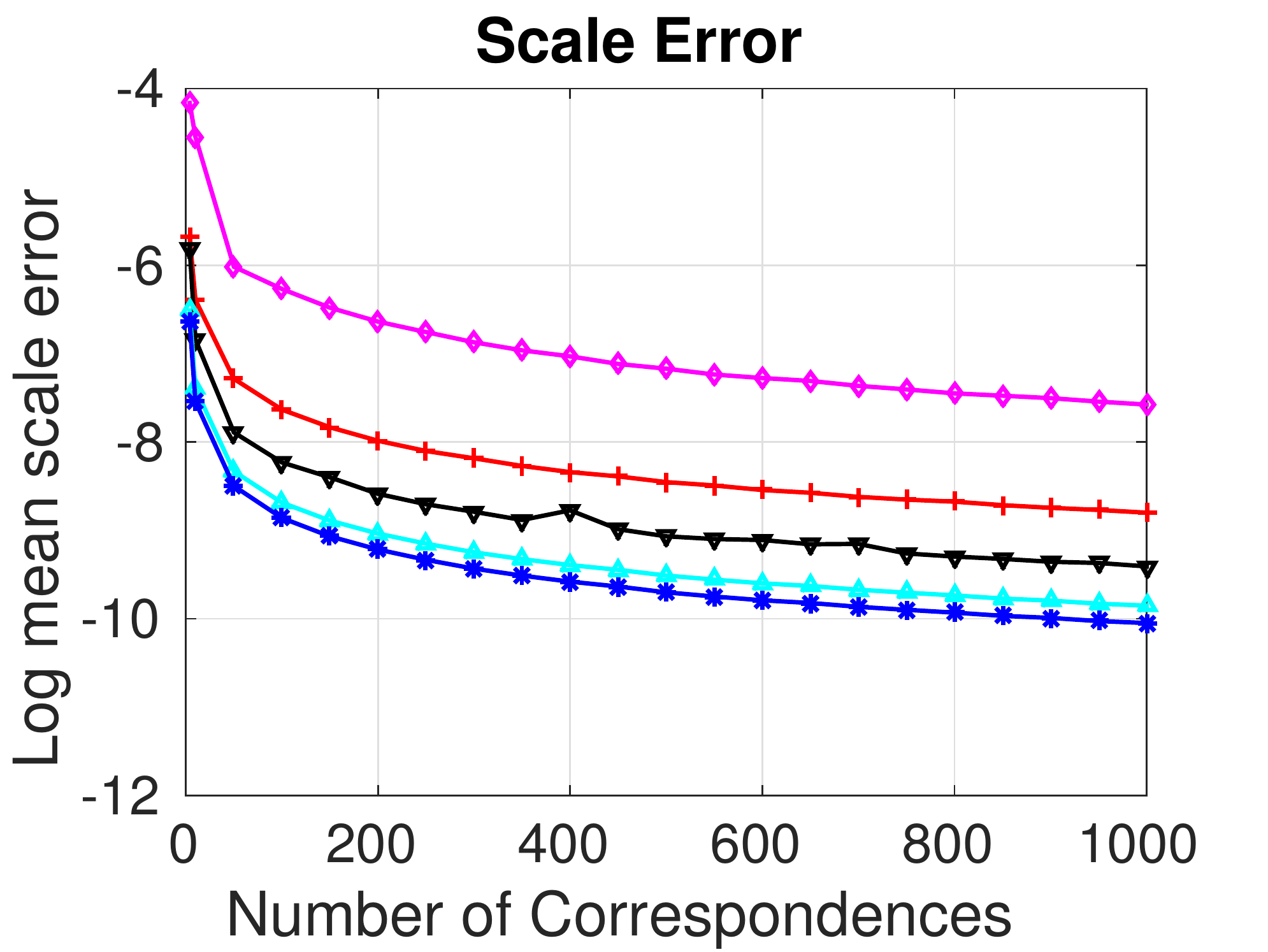}
\end{center}
\vspace{-0.1in}
\caption{\label{fig:more_correspondences} 
We measured the accuracy of similarity transformation estimations as the number of correspondences increased. The mean of the log rotation, translation, and scale errors are plotted from 1000 trials at each level of correspondences used. A Gaussian image noise of 0.5 pixels was used for all trials. We did not use P3P+s in this experiment because P3P only uses 3 correspondences. Our algorithm has better accuracy for all number of correspondences used and a runtime complexity of $O(n)$, making it ideal for use at scale.
}
\end{figure}
\textbf{Scalability experiment:} For the second experiment, we evaluate the similarity transformation error as the number of 2D-3D correspondences increases. We use the same camera configuration described above, but vary the number of 3D points used to compute the similarity transformation from 4 to 1000. We ran 1000 trials for each number of correspondences used with a Gaussian noise level of 0.5 pixels standard deviation for all trials. We did not use the P3P+s algorithm for this experiment since P3P is a minimal solver and cannot utilize the additional correspondences. The accuracy of each similarity transformation algorithm as the number of correspondences increases is shown in Figure~\ref{fig:more_correspondences}. Our algorithm performs very well as the number of correspondences increases, and is more accurate than alternative algorithms for all numbers of correspondences tested. Further, our algorithm is $O(n)$ so the performance cost of using additional correspondences is favorable compared to the alternative algorithms.

\subsection{SLAM Registration With Real Images}
We tested our new solver for registration of a SLAM reconstruction with respect to an existing SfM reconstruction using the indoor dataset from~\cite{ventura2014cvpr}. This dataset consists of an indoor reconstruction with precise 3D and camera position data obtained with an ART-2 optical tracker. All algorithms are used in a PROSAC~\cite{chum2005matching} loop to estimate similarity transformations from 2D-3D correspondences. We compare these algorithms to our gDLS+++ algorithm. 

\begin{table*}[t]
\begin{flushleft}
   \caption{ \label{slam_error} Average position error in centimeters for aligning a SLAM sequence to a pre-existing SfM reconstruction. An ART-2 tracker was used to provide highly accurate ground truth measurements for error analysis. Camera positions were computed using the respective similarity transformations and the mean camera position error of each sequence is listed below. Our method, gDLS+++, outperforms all other methods and is extremely close to the solution after BA.}
   \vspace{-0.2in}
   \end{flushleft}
      \centering
       \begin{tabular}{| c | c || r | r | r | r | r || r | r |}
    \hline
    Sequence & \# Images & Abs. Ori.~\cite{umeyama1991least} & P3P+s & P$n$P+s & gP+s\cite{ventura2014cvpr} & gDLS~\cite{sweeney2014gdls} & gDLS+++ & After BA \\ \hline 
   office1   & 9  & 6.37 & 6.14 & 4.38 & 6.12 & 3.97 & 3.68 & 3.61 \\  
   office2   & 9  & 8.09 & 7.81 & 6.90 & 9.32 & 5.89 & 5.59 & 5.57 \\  
   office3   & 33 & 8.29 & 9.31 & 8.89 & 6.78 & 6.08 & 4.91 & 4.86 \\  
   office4   & 9  & 4.76 & 4.48 & 3.98 & 4.00 & 3.81 & 3.09 & 3.04 \\  
   office5   & 15 & 3.63 & 3.42 & 3.39 & 4.75 & 3.39 & 3.17 & 3.14 \\  
   office6   & 24 & 5.15 & 5.23 & 5.01 & 5.91 & 4.51 & 4.35 & 4.31 \\  
   office7   & 9  & 6.33 & 7.08 & 7.16 & 7.07 & 4.65 & 2.99 & 2.72 \\  
   office8   & 11 & 4.72 & 4.85 & 3.62 & 4.59 & 2.85 & 2.30 & 2.12 \\  
   office9   & 7  & 8.41 & 8.44 & 4.08 & 6.65 & 3.19 & 2.25 & 2.25 \\  
   office10  & 23 & 5.88 & 6.60 & 5.73 & 5.88 & 4.94 & 4.68 & 4.61 \\  
   office11  & 58 & 5.19 & 4.85 & 4.80 & 6.74 & 4.77 & 4.66 & 4.57 \\  
   office12  & 67 & 5.53 & 5.20 & 4.97 & 4.86 & 4.81 & 4.45 & 4.44 \\  
   \hline
    \end{tabular}
    \vspace{-0.1in}
\end{table*}

We compute the average position error of all keyframes with respect to the ground truth data. The position errors, reported in centimeters, are shown in Table~\ref{slam_error}. Our gDLS+++ solver gives higher accuracy results for every image sequence tested compared to alternative algorithms. By using the generalized camera model, we are able to exploit 2D-3D constraints from multiple cameras at the same time as opposed to considering only one camera (such as P3P+s and P$n$P+s). This allows the similarity transformation to be optimized for all cameras and observations simultaneously, leading to high-accuracy results. 

We additionally show the results of gDLS+++ with bundle adjustment applied after estimation (labeled ``After BA" in Table~\ref{slam_error}). In all datasets, our results are very close to the optimal results after bundle adjustment, and typically bundle adjustment converges after only one or two iterations. This indicates that the gDLS+++ algorithm is very close to the geometrically optimal solution in terms of reprojection error. Further, our singularity-free rotation parameterization prevents numerical instabilities that arise as the computed rotation approaches a singularity, leading to more accurate results than the gDLS~\cite{sweeney2014gdls} algorithm. 

\section{Incremental SfM with Distributed Cameras}
\label{sec:sfm}

\begin{table*}[t]
\begin{flushleft}
   \caption{ \label{fig:sfm_results} Results of our Hierarchical SfM pipeline on several large scale datasets. Our method is extremely efficient and is able to reconstruct more cameras than the Divide-and-Conquer method of Bhowmick \etal~\cite{divide2014accv}. Time is provided in minutes.}
   \vspace{-0.2in}
   \end{flushleft}
   \centering
       \begin{tabular}{| l | c || c | c | c | c | c | c | c | c | c |}
    \hline
    \multirow{2}{*}{Dataset}& \multirow{2}{*}{$N_{cam}$}& \multicolumn{2}{|c|}{Visual SfM~\cite{wu2013towards}} & \multicolumn{2}{|c|}{DISCO~\cite{disco2013pami}} & \multicolumn{2}{|c|}{Bhowmick \etal~\cite{divide2014accv}} &   \multicolumn{3}{|c|}{Ours} \\ \cline{3-11}
     &  & $N_{cam}$ & Time & $N_{cam}$ & Time & $N_{cam}$ & Time & $N_{cam}$ & BA Its & Time \\ \hline 
 Colosseum & 1164 & 1157 & 9.85 & N/A & N/A & 1032 & 3 & 1097 & 1 & 2.6 \\
 St Peter's Basilica & 1275 & 1267 & 9.71 & N/A & N/A & 1236 & 4 & 1256 & 1 & 3.7 \\
 Dubrovnik & 6845 & 6781 & 16.8 & 6532 & 275 & N/A & N/A  & 6677 & 2 & 8.9 \\
 Rome & 15242 & 15065 & 100.17 & 14754 & 792 & 10534 & 27 &  12329 & 2 & 22\\
   \hline
    \end{tabular}
    \vspace{-0.1in}
\end{table*}

\begin{figure}[t]
\vspace{-0.15in}
\centering
\includegraphics[width=0.9\linewidth,keepaspectratio]{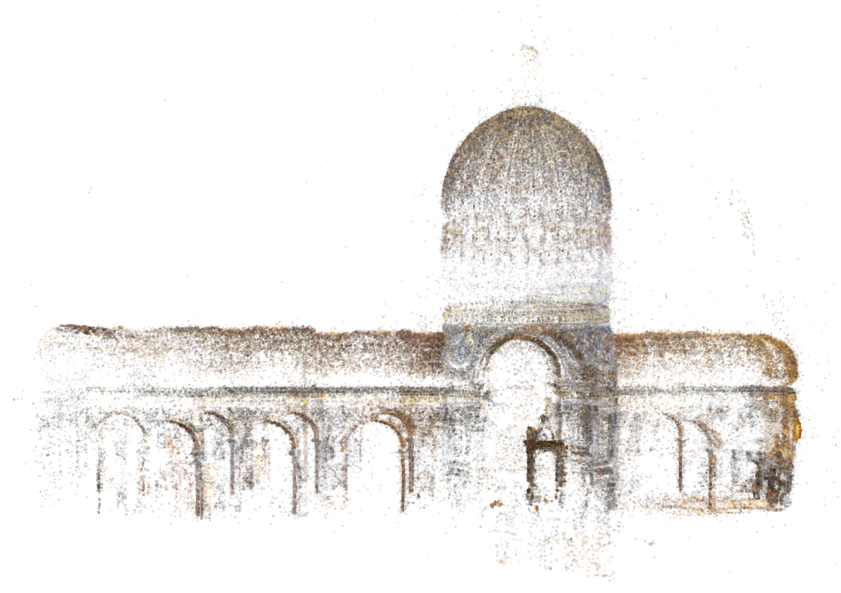}
\vspace{-0.1in}
\includegraphics[width=0.9\linewidth,keepaspectratio]{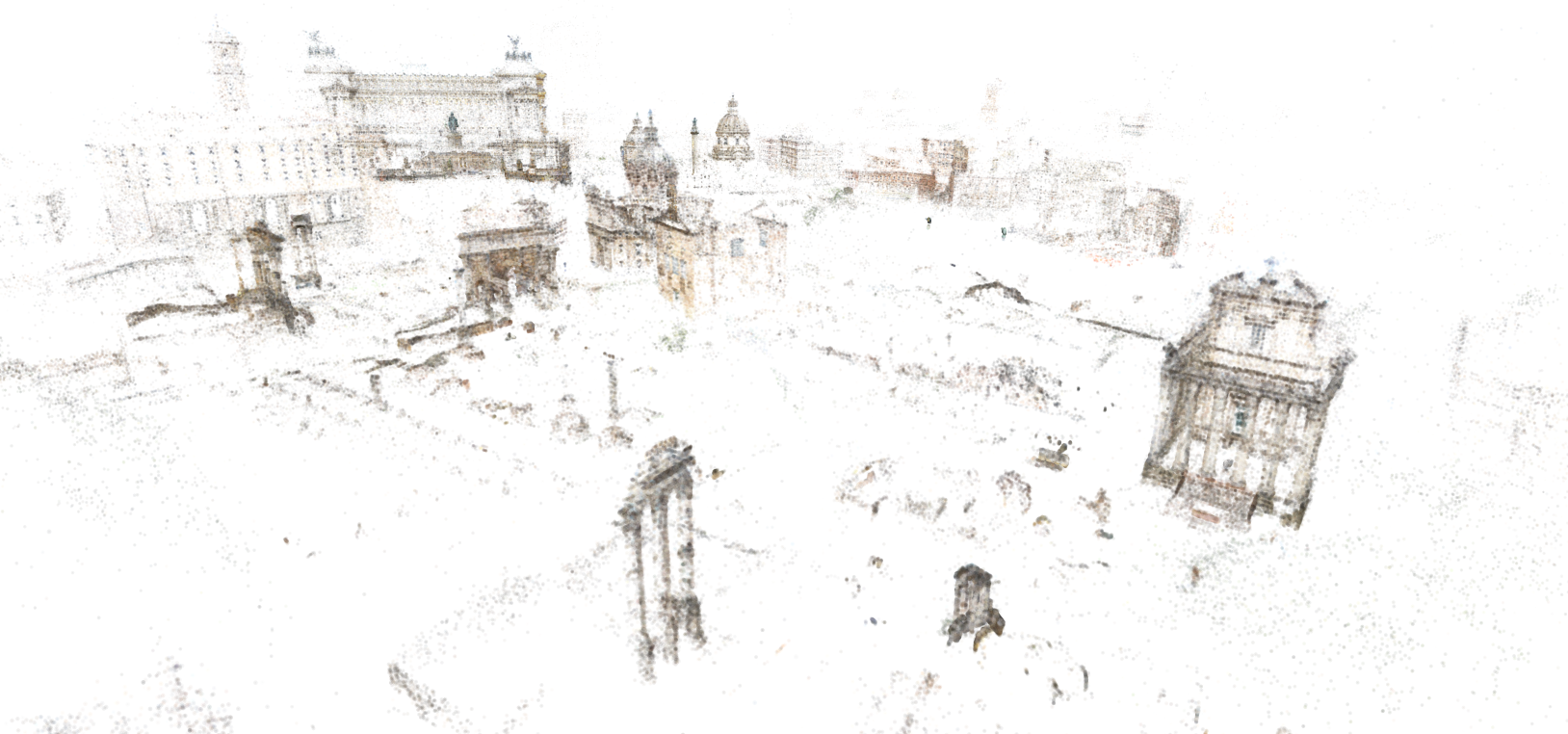}
\caption{
\label{fig:merged_models} 
Final reconstructions of the St Peters (top) and Central Rome (bottom) datasets computed with our hierarchical SfM pipeline. Our pipeline produces high quality visual results at state-of-the-art efficiency (\cf Table~\ref{fig:sfm_results}).
}
\vspace{-0.1in}
\end{figure}

We demonstrate the viability of our method for SfM model-merging in a novel incremental SfM pipeline. Our pipeline uses distributed cameras to rapidly grow the model by adding many cameras at once. We use the gDLS+++ method to localize a distributed camera to our current model in the same way that traditional incremental pipelines use P3P to localize individual cameras. This allows our pipeline to be extremely scalable and efficient because we can accurately localize many cameras to our model in a single step. Note that a 3D reconstruction of points and cameras may be alternatively described as a distributed camera where ray origins are the camera origins and the ray directions (and depths) correspond to observations of 3D points. In our incremental SfM procedure we treat reconstructions as distributed cameras, allowing reconstructions to be merged efficiently and accurately with the gDLS+++ algorithm.

We begin by partitioning the input dataset into subsets using normalized cuts~\cite{shi2000normalized} similar to Bhowmick ~\etal~\cite{divide2014accv}. The size of the subsets depends on the size of the particular dataset, but typically the subsets contain between 50 and 250 cameras. Each of the subsets is individually reconstructed in parallel with the ``standard" incremental SfM pipeline of the Theia SfM library~\cite{theia-manual}. Each reconstructed subset may be viewed as a distributed camera. The remaining task is then to localize all distributed cameras into a common coordinate system in the same way that traditional incremental SfM localizes individual cameras as it grows the reconstruction.

The localization step operates in a similar manner to the first step. Distributed cameras are split into subsets with normalized cuts~\cite{shi2000normalized}. Within each subset of distributed cameras, the distributed camera with the most 3D points is chosen as the ``base" and all other distributed cameras are localized to the ``base" distributed camera with the gDLS+++ algorithm. Note that when two distributed cameras are merged (with gDLS or another algorithm) the result is another distributed camera which contains all observation rays from the distributed cameras that were merged. As such, each merged subset forms a new distributed camera that contains contains all imagery and geometric information of the cameras in that subset. This process is repeated until only a single distributed camera remains (or no more distributed cameras can be localized). Note that by using distributed cameras we not only increase the $\Delta$ at which we can grow models, but $\Delta$ grows at an exponential rate because larger and larger distributed cameras are added to the model as the reconstruction process continues. We do not run bundle adjustment as subsets are merged and only run a single global bundle adjustment on the entire reconstruction as a final refinement step. Avoiding the use of costly bundle adjustment is a driving reason for why our method is very efficient and scalable.

We compare our SfM pipeline to Incremental SfM (VisualSfM~\cite{wu2013towards}), the DISCO pipeline of Crandall \etal~\cite{disco2013pami}, and the hierarchical SfM pipeline of Bhowmick \etal~\cite{divide2014accv} run on several large-scale SfM datasets and show the results in Table ~\ref{fig:sfm_results}. All methods were run on a desktop machine with a 3.4GHz i7 processor and 16GB of RAM. Our method is the most efficient on all datasets, though we typically reconstruct fewer cameras than ~\cite{wu2013towards}. Using gDLS+++ for model-merging allows our method produces high quality models by avoiding repeated use of bundle adjustment. As shown in Table ~\ref{fig:sfm_results}, the final bundle adjustment for our pipeline requires no more than 2 iterations, indicating that the gDLS+++ method is able to merge reconstructions extremely accurately. We show the high quality visual results of our SfM pipeline in Figure ~\ref{fig:merged_models}.

\section{Conclusion}
\label{sec:conclusion}
In this paper, we introduced a new camera model for SfM: the distributed camera model. This model describes image observations in terms of light rays with ray origins and directions rather than pixels. As such, the distributed camera model is able to describe a single camera or multiple cameras in a unified expression as the collection of all light rays observed. We showed how the gDLS method~\cite{sweeney2014gdls} is in fact a generalization of the standard absolute pose problem and derive an improved solution method, gDLS+++, that is able to solve the absolute pose problem for standard or distributed cameras. Finally, we showed how gDLS+++ can be used in a scalable incremental SfM pipeline that uses the distributed camera model to accurately localize many cameras to a reconstruction in a single step. As a result, fewer bundle adjustments are performed and the resulting efficient pipeline is able to reconstruct a 3D model of Rome from more than 15,000 images in just 22 minutes. We believe that the distributed camera model is a useful way to parameterize cameras and can provide great benefits for SfM. For future work, we plan to explore the use of distributed cameras for global SfM, as well as structure-less SfM by merging distributed cameras from 2D-2D ray correspondences without the need for 3D points~\cite{kneip2016generalized, sweeney2015computing}.


\section*{Acknowledgments}

The authors would like to thank Jonathan Ventura for his insights and for providing the SLAM datasets used in our real data experiments. 
This work was supported by NSF grant IIS-1423676 and ONR grant N00014-14-1-0133.

{\small
\bibliographystyle{ieee}
\bibliography{gdls_hierarchical_sfm}
}

\end{document}